\documentclass{article}

\usepackage[main, final]{neurips_2026}
\usepackage{tabularx}
\usepackage{amsmath}
\usepackage{hyperref}
\usepackage{graphicx}
\usepackage{cleveref}
\usepackage{float}
\usepackage{multirow}

\usepackage[utf8]{inputenc} 
\usepackage[T1]{fontenc}    
\usepackage{url}            
\usepackage{booktabs}       
\usepackage{amsfonts}       
\usepackage{nicefrac}       
\usepackage{microtype}      
\usepackage{xcolor}         
\usepackage{wrapfig}
\usepackage[table]{xcolor}

\definecolor{best}{RGB}{244,170,170}    
\definecolor{second}{RGB}{248,208,164}  
\definecolor{third}{RGB}{255,246,170}   

\newcommand{\best}[1]{\cellcolor{best}#1}
\newcommand{\second}[1]{\cellcolor{second}#1}
\newcommand{\third}[1]{\cellcolor{third}#1}

\title{TransmissiveGS: Residual-Guided Disentangled Gaussian Splatting for Transmissive Scene Reconstruction and Rendering}

\author{%
  \textbf{Zhenyu Liang} \quad
  \textbf{Xiao Zhang} \quad
  \textbf{Tianchao Li} \quad
  \textbf{Jack C.P. Cheng} \quad
  \textbf{Chi-Keung Tang} \\[0.5em]
  \textnormal{The Hong Kong University of Science and Technology}
}

\begin{document}

\maketitle

\begin{figure}[h]

  \centering
  \includegraphics[width=\textwidth]{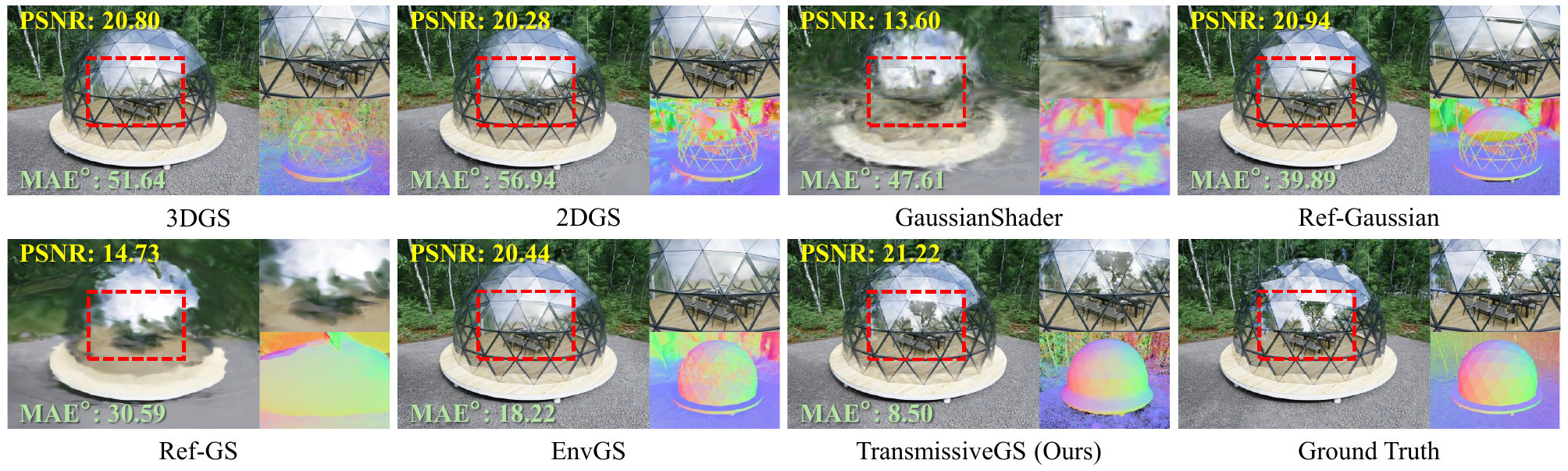}

\caption{{\bf Comparison on a challenging example consisting of a hemispherical, reflective, and transmissive dome.} We present {\bf TransmissiveGS}, the 
\textit{first} Gaussian Splatting framework that achieves both accurate transmissive 
surface reconstruction (MAE$^\circ$ $\downarrow$) and photorealistic inverse rendering (PSNR $\uparrow$) in non-planar transmissive scenes featuring near-field reflections and visible transmitted content behind the transmissive surface, without requiring any geometry prior.}
  \label{fig:opening}

\end{figure}

\begin{abstract}
   Transmissive scenes are ubiquitous in daily life, yet reconstructing and rendering them remains highly challenging due to the inherent entanglement between near-field reflections from the surrounding environment on the transmissive surface, and the transmitted content of the scene behind it. This coupling gives rise to dual surface geometries and dual radiance components within each observation, posing ambiguities for standard methods. We present TransmissiveGS, a novel framework for disentangled reconstruction and rendering of transmissive scenes. Specifically, we model the scene with a dual-Gaussian representation and introduce a deferred shading function to jointly render the two Gaussian components. To separate reflection and transmission, we exploit the inherent multi-view inconsistency of reflections and leverage the residuals from reconstructing multi-view consistent content as cues for disentangled geometry and appearance modeling. We further propose a reflection light field that enables high-fidelity estimation of near-field reflections. During training, we introduce a high-frequency regularization to preserve fine details. We also contribute a new synthetic dataset for evaluating transmissive surface reconstruction. Experiments on both synthetic and real-world scenes demonstrate that TransmissiveGS consistently outperforms prior Gaussian Splatting-based methods in both reconstruction and rendering quality for transmissive scenes.
\end{abstract}


\section{Introduction}
\label{sec:intro}

Accurately reconstructing complex scenes from multi-view captures with photorealistic inverse rendering has long been a central problem in computer vision and computer graphics (Figure~\ref{fig:opening}). Recently, neural radiance fields (NeRF~\cite{mildenhall2020nerf}), based on implicit multi-layer perceptrons (MLP), and Gaussian Splatting (GS~\cite{kerbl20233d}), based on explicit Gaussian primitives, have brought substantial breakthroughs. Compared to NeRF, GS leverages rasterization to enable fast training and high-fidelity rendering. Although several recent works~\cite{yao2025reflective,jiang2024gaussianshader,zhang2025ref,xie2025envgs,ye20243d,gu2025irgs,zhangmaterialrefgs} have extended GS to handle reflective surfaces, the modeling of transmissive surfaces remains largely unexplored.

Objects with transmissive surfaces, such as glass, are challenging because their observed appearance is a mixture of two components: the reflected radiance from the transmissive surface itself, and the transmitted radiance from surfaces behind it. As a result, the observations in captured images entangle information from two distinct geometric surfaces. Most prior work represents the scene with a single set of Gaussians, which introduces ambiguity when reconstructing transmissive surfaces. Specifically, since the transmitted component is typically multi-view consistent, whereas the reflected component is high-frequency and strongly view-dependent, existing methods tend to explain the observations by fitting the transmitted surfaces, causing the transmissive surfaces to be poorly reconstructed. This, in turn, degrades the estimation and rendering of reflections. 




Meanwhile, in real-world scenarios, the reflected radiance is often dominated by high-frequency near-field illumination. Vanilla GS methods~\cite{kerbl20233d,huang20242d} represent color using spherical harmonics (SH), which can only capture low-frequency view-dependent appearance. Some works~\cite{yao2025reflective,jiang2024gaussianshader,ye20243d,gu2025irgs,zhangmaterialrefgs} instead use an environment map to model global illumination. While this approach can partially address high-frequency view-dependent effects, the environment map assumption that illumination originates from infinity introduces parallax errors under near-field lighting. EnvGS~\cite{xie2025envgs} addresses this limitation by explicitly reconstructing the surrounding environment geometry as environment Gaussians to represent near-field illumination, followed by querying them via ray tracing. However, when surface reflections are weak, as is typical in transmissive 
scenes, the reconstructed environment Gaussians lack sufficient supervisory 
signal, leading to degraded quality. Ref-GS~\cite{zhang2025ref} leverages surface light fields by augmenting directional encoding with spatial features and predicting reflected color via a lightweight MLP; nevertheless, its capacity for high-frequency details remains limited.

\begin{figure}[b]

  \centering
  \includegraphics[width=\textwidth]{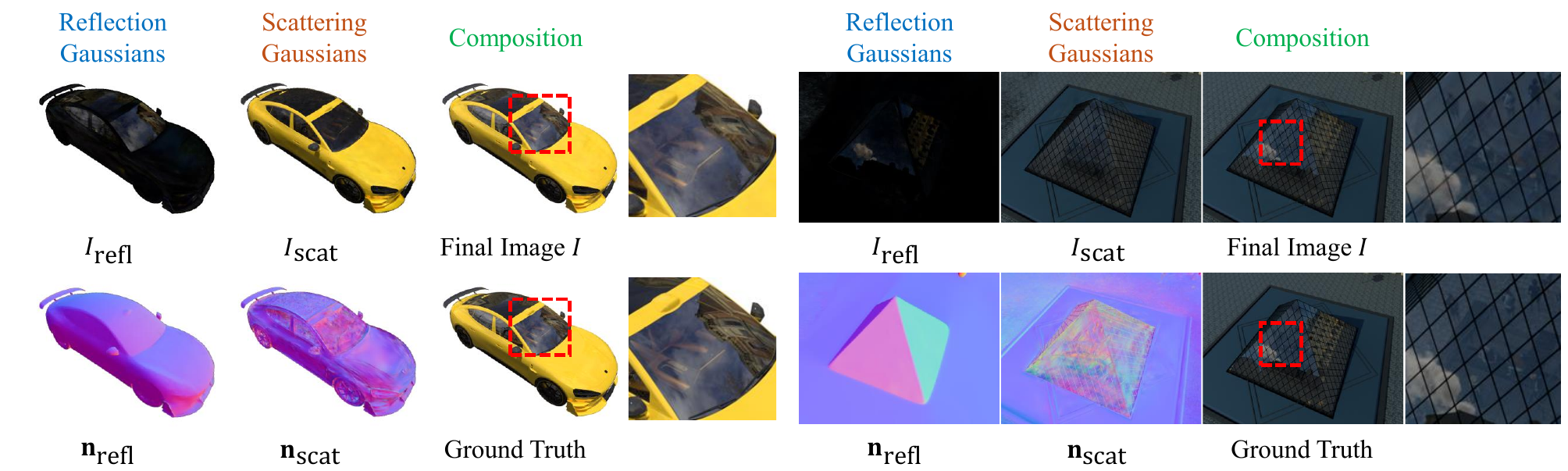}

\caption{{\bf Disentanglement and composite appearance.} \textit{Left:} In the \textit{car} case, the windshield exhibits both 
reflections of the surrounding environment and transmitted interior objects. 
\textit{Right:} In the \textit{pyramid} case, the glass surface reflects near-field 
surrounding structures while transmitting the people inside. Please see text for symbol definitions.}
  \label{fig:disganlement}
  \vspace{-0.2in}
\end{figure}

To overcome these challenges, we present {\bf TransmissiveGS}, which (i) effectively disentangles and reconstructs the geometries of both transmissive surfaces and the transmitted surfaces visible behind them, and (ii) faithfully models high-frequency, near-field reflected radiance along with its composition with transmitted radiance (Figure~\ref{fig:disganlement}). To this end, we introduce a {\bf dual-Gaussian} scene representation that decomposes the scene into reflection Gaussians, capturing reflective surfaces including the transmissive surfaces themselves, and scattering Gaussians, capturing the remaining scene content visible through or around them. A novel deferred shading function composites the rendered outputs of both Gaussian sets to render the scenes. To resolve the geometric ambiguity of transmissive surfaces, we propose a {\bf residual-guided disentangling} strategy: we first reconstruct the multi-view consistent content via the scattering Gaussian, then exploit the resulting {\bf color residuals} to supervise the reflection Gaussian in recovering the transmissive surface geometry. To faithfully reproduce the high-frequency, near-field reflections observed on these surfaces, we further introduce a {\bf reflection light field} conditioned on encoded surface position and reflection direction. Training additionally incorporates a high-frequency regularization to preserve fine details.

To summarize, our contributions are as follows:
\vspace{-3mm}
\begin{itemize}
\item A dual-Gaussian scene representation with a deferred shading function that jointly models reflected and transmitted radiance and their composite appearance in transmissive scenes.
\vspace{-1mm}
\item A residual-guided reconstruction strategy that leverages multi-view fitting residuals to disentangle and recover transmissive surface geometry.
\vspace{-1mm}
\item A reflection light field that estimates high-frequency, near-field reflections, conditioned on surface position and reflection direction.
\end{itemize}
\vspace{-3mm}

We additionally contribute a new synthetic dataset specifically designed for evaluating transmissive surface reconstruction, addressing the absence of such benchmarks in existing public datasets~\cite{liu2023nero,verbin2024ref}. Extensive experiments on both synthetic and real-world scenes demonstrate that TransmissiveGS consistently outperforms state-of-the-art GS-based methods.

\section{Related Work}

\paragraph{Inverse Rendering.}
Inverse rendering recovers physical scene properties (e.g., geometry, 
materials, and lighting) from observed images and re-renders the scene 
based on the recovered attributes~\cite{liang2024gs,boss2021nerd}. Early 
approaches build on NeRF~\cite{mildenhall2020nerf}: 
NeRFactor~\cite{zhang2021nerfactor} and PhySG~\cite{zhang2021physg} 
factorize shape, spatially-varying BRDF, and illumination; 
NeRO~\cite{liu2023nero} and NeILF++~\cite{zhang2023neilf++} introduce 
incident light fields to handle inter-reflections; and 
Ref-NeRF~\cite{verbin2024ref} improves specular reconstruction through 
structured view-dependent parameterizations. Despite their effectiveness, 
NeRF-based methods incur substantial computational overhead due to dense 
per-sample MLP evaluations.
Gaussian Splatting (GS)~\cite{kerbl20233d} offers an efficient alternative 
through explicit primitives and differentiable rasterization. Recent works 
extend GS to inverse rendering along several axes: \emph{forward shading} 
methods~\cite{jiang2024gaussianshader,liang2024gs} evaluate per-Gaussian 
shading before alpha-blending but suffer from specular noise; \emph{deferred 
shading} approaches~\cite{ye20243d,zhou2025rtr} rasterize attributes into 
screen-space buffers for more stable optimization of glossy surfaces; 
\emph{near-field illumination} methods~\cite{tang2025spectre,gu2025irgs,
xie2025envgs} go beyond distant environment maps by reconstructing the 
surrounding light field and querying it via ray tracing; and \emph{implicit} 
approaches~\cite{zhang2025ref,tang20243igs} predict reflected radiance via 
MLPs for greater flexibility. A few works~\cite{huang2025transparentgs,
zhang2025ref} further handle \emph{refraction}, but target fully transparent 
objects without interior visible surfaces. More recent efforts on coupled \emph{reflective} and \emph{transmissive} scenes, such as TR-Gaussians~\cite{liu2026tr} and 
RT-GS~\cite{zeng2026rt}, are restricted to specific planar or 
semi-transparent surfaces and do not generalize to non-planar or fully 
transparent ones. TSGS~\cite{li2025tsgsb} focuses only on transparent objects without reflections. Overall, inverse rendering through transmissive surfaces 
remains largely underexplored.

\paragraph{Disentanglement of Reflection and Transmission.}
Separating reflected and transmitted components through transmissive surfaces 
is a long-standing challenge in both 2D image 
processing~\cite{levin2007user,li2014single,shih2015reflection,
fan2017generic,wei2019single,li2020single,lei2020polarized,
xue2015computational,guo2014robust} and 3D scene reconstruction. 
NeRFReN~\cite{guo2022nerfren} pioneered this decomposition by modeling 
separate transmitted and reflected radiance fields, but treats reflections as 
a view-independent virtual image, an assumption valid only for planar 
reflectors. MS-NeRF~\cite{yin2023multi} and NeuS-HSR~\cite{qiu2023looking} 
mitigate reflection-induced artifacts via parallel sub-space features and 
SDF-based auxiliary planes, respectively, yet neither imposes explicit 
decomposition constraints nor reconstructs the transmissive surface geometry. 
RA-NeRF~\cite{gao2024planar} improves physical plausibility by casting 
reflected rays through a jointly optimized planar reflector, but remains 
restricted to planar geometry. Within the Gaussian Splatting paradigm, 
Flash-Splat~\cite{xie2024flash} achieves high-quality disentanglement using 
unpaired flash/no-flash cues, though it requires controlled illumination that 
limits in-the-wild applicability. DC-Gaussian~\cite{wang2024dc} uses image decomposition to separate windshield reflections and occlusions in driving scenes. GlassGaussian~\cite{cao2025glassgaussian} models directional reflectance, while TR-Gaussians~\cite{liu2026tr} combines mirrored Gaussians with Fresnel blending. Both methods assume planar glass geometry. In summary, existing methods 
either rely on planar surface assumptions or fail to accurately reconstruct 
the transmissive surface geometry itself. Our TransmissiveGS addresses these 
limitations, enabling the joint recovery of geometry, materials, and lighting 
for general transmissive scenes, including those with curved surfaces.

\section{Preliminary}
\label{sec:preli}

Our TransmissiveGS builds upon 2D Gaussian Splatting (2DGS~\cite{huang20242d}) as the underlying scene primitive for reconstruction and rendering.
Unlike 3D Gaussian Splatting (3DGS~\cite{kerbl20233d}), which models a scene with volumetric Gaussian ellipsoids and is prone to multi-view inconsistency in surface normal estimation, 2DGS employs flat Gaussian elliptical disks that are intrinsically aligned with scene surfaces, thereby yielding substantially higher geometric reconstruction quality.

Each 2D Gaussian disk is parameterized by its center position $\mathbf{p}_k$, two principal tangent vectors $\mathbf{t}_u$ and $\mathbf{t}_v$ with associated scaling factors $s_u$ and $s_v$, an opacity value $\alpha$, and a set of spherical harmonics (SH) coefficients $\mathbf{c}$ encoding view-dependent color.
A point on the disk is expressed in world space through a local tangent-plane parameterization:
\begin{equation}
    P(u,v) = \mathbf{p}_k + s_u \mathbf{t}_u\, u + s_v \mathbf{t}_v\, v
            = H\,(u,\,v,\,1,\,1)^{\top}
\end{equation}
where $ H =
    \begin{bmatrix}
        s_u\mathbf{t}_u & s_v\mathbf{t}_v & \mathbf{0} & \mathbf{p}_k \\
        0 & 0 & 0 & 1
    \end{bmatrix}$
is a homogeneous transformation matrix that encodes the geometry of the disk.
Given a local coordinate $\mathbf{u}=(u,v)$ on the tangent plane, the Gaussian evaluation is defined as:
\begin{equation}
    \mathcal{G}(\mathbf{u})
    = \exp\!\left(-\frac{u^2 + v^2}{2}\right)
\end{equation}

During rendering, each image pixel $\mathbf{x}=(x,y)$ is mapped to the local coordinate of a given disk via a ray--splat intersection.
Letting $W\in\mathbb{R}^{4\times 4}$ denote the world-to-screen transformation matrix and $z$ the depth at the intersection point, this mapping satisfies:
\begin{equation}
    \mathbf{x} = (xz,\; yz,\; z,\; 1)^{\top}
               = W\, H\,(u,\,v,\,1,\,1)^{\top}
    \label{eq:eq3}
\end{equation}

The final pixel color is then composited via front-to-back alpha blending over the sorted splats:
\begin{equation}
    \mathbf{c}(\mathbf{x}) =
    \sum_{i=1}^{N} \mathbf{c}_i\, \alpha_i\, \hat{\mathcal{G}}_i\!\bigl(\mathbf{u}(\mathbf{x})\bigr)
    \prod_{j=1}^{i-1}
    \Bigl(1 - \alpha_j\, \hat{\mathcal{G}}_j\!\bigl(\mathbf{u}(\mathbf{x})\bigr)\Bigr)
    \label{eq:eq4}
\end{equation}
where $\hat{\mathcal{G}}$ is a low-pass filtered variant of the Gaussian kernel, introduced to suppress numerical instability under highly oblique viewing angles~\cite{huang20242d}.

\section{TransmissiveGS}
\label{sec:method}

\begin{figure}[t]

  \centering
  \includegraphics[width=\textwidth]{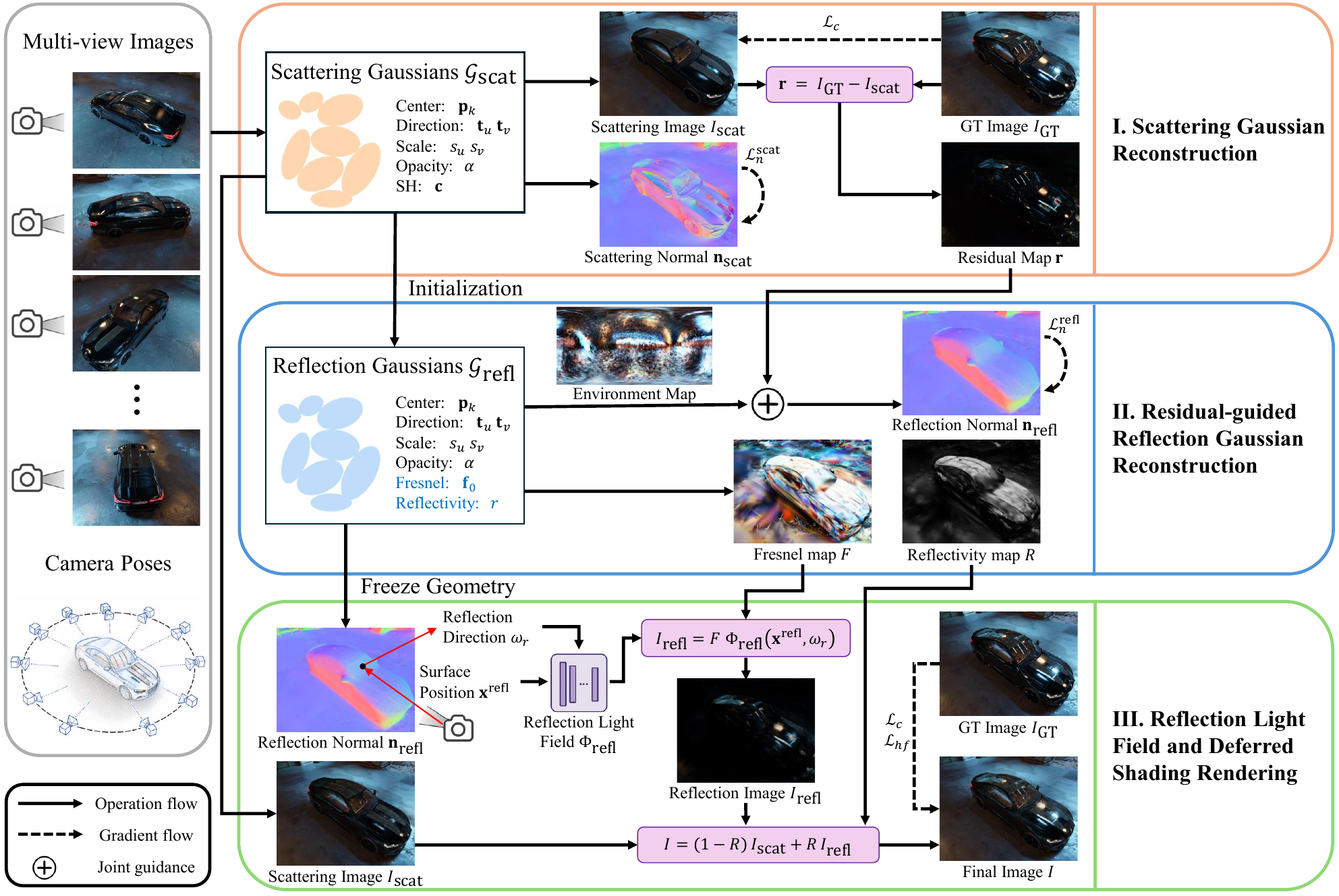}

\caption{{\bf Pipeline of our proposed TransmissiveGS.}
Our framework consists of three stages.
In \textbf{Stage~I}, the scattering Gaussians are trained to reconstruct multi-view consistent scene content.
In \textbf{Stage~II}, the residual signal, together with an environment map, jointly supervises the reflection Gaussians to recover the geometry of reflective and transmissive surfaces.
In \textbf{Stage~III}, a reflection light field is trained to capture near-field, view-dependent reflections, and the final image is produced via a deferred shading composition of the scattering image and the reflection image.
}
  \label{fig:TransmissiveGS}

\end{figure}

We propose TransmissiveGS to render photorealistic transmissive scenes
while faithfully reconstructing their geometry.
Figure~\ref{fig:TransmissiveGS} provides an overview of the pipeline.

\subsection{Dual-GS Representation and Deferred Shading Function}

When light encounters a transmissive surface, part of the incident
energy is reflected while the remainder is transmitted and subsequently
scattered by the interior structure.  Motivated by this physical
observation, we employ a \emph{dual} Gaussian representation that
disentangles the two transport modes:
(i)~a set of \emph{reflection Gaussians}
$\mathcal{G}_{\mathrm{refl}}$ that model the reflective surfaces,
including the transmissive surfaces themselves, and capture the
high-frequency, view-dependent reflections; and
(ii)~a set of \emph{scattering Gaussians}
$\mathcal{G}_{\mathrm{scat}}$ that represent the remaining scene
content visible through or around the transmissive surfaces, capturing
the multi-view consistent diffuse radiance.
The two sets are rasterized independently, and the resulting images are
merged in screen space via a physically motivated deferred shading
function.

The scattering Gaussians follow the standard 2DGS parameterization
described in Section~\ref{sec:preli}:  Rasterizing them produces a per-pixel
scattering image~$I_{\mathrm{scat}}$.
The reflection Gaussians augment this base parameterization with two
additional learnable attributes: a Fresnel vector
$\mathbf{f}_0\!\in\mathbb{R}^{3}$ that controls the
wavelength-dependent base reflectance, and a scalar reflectivity
coefficient $r\!\in\mathbb{R}$ that encodes the overall proportion of
reflected energy at each Gaussian.

After rasterization, the two components are composited on a per-pixel
basis.  Let $R$ denote the reflectivity map obtained by
rasterizing~$r$. The final pixel color is given by:
%
\begin{equation}
I
= \bigl(1-R\bigr)\,I_{\mathrm{scat}}
+ R\,I_{\mathrm{refl}}
\label{eq:eq6}
\end{equation}
where the first term accounts for the transmitted and scattered light
and the second term captures the surface reflection.

The reflection image~$I_{\mathrm{refl}}$ is obtained by querying
the reflection light field~$\Phi_{\mathrm{refl}}$ (described in
Section~\ref{sec:reflmlp}) at the corresponding surface point
$\mathbf{x}^{\mathrm{refl}}\!\in\mathbb{R}^{3}$ on the reflection
Gaussians, and modulating the returned environment radiance with a
Fresnel term~$F$ to account for view-dependent reflectance:
\begin{equation}
I_{\mathrm{refl}}
= F \;\Phi_{\mathrm{refl}}\!\bigl(\mathbf{x}^{\mathrm{refl}},\omega_r\bigr)
\label{eq:eq7}
\end{equation}


%
Here, the reflection direction~$\omega_r$ is computed from the
viewing direction~$\omega_o$ (from camera to
surface) and the surface
normal~$\mathbf{n}_{\mathrm{refl}}$ as:
\begin{equation}
\omega_r =  \omega_o - 2(\omega_o^{\top}\mathbf{n}_{\mathrm{refl}})\,\mathbf{n}_{\mathrm{refl}}
\label{eq:eq8}
\end{equation}
Assuming ideal specular reflection, the Fresnel term~$F$ is evaluated
with the Schlick approximation:
\begin{equation}
  F
  = F_0
  + \bigl(1-F_0\bigr)
    \bigl(1-\mathbf{n}_{\mathrm{refl}}^{\top}(-\omega_o)\bigr)^{5}
  \label{eq:eq9}
\end{equation}
where $F_0$ is a per-pixel Fresnel coefficient map obtained by
rasterizing the learnable Fresnel vector~$\mathbf{f}_0$ of the
reflection Gaussians.

\subsection{Residual-Guided Disentangling Reconstruction}
\label{sec:residual}

As shown in Figure~\ref{fig:TransmissiveGS}, our framework consists
of three stages. In Stage~I, we optimize
only the scattering Gaussians $\mathcal{G}_{\mathrm{scat}}$ from
multi-view images.  Because diffuse surfaces exhibit strong multi-view
photometric consistency, $\mathcal{G}_{\mathrm{scat}}$ can reconstruct
them faithfully.  Reflective and transmissive surfaces, however,
introduce view-dependent appearance,
causing the optimizer to leave these regions unoccupied.  A per-pixel photometric residual map $\mathbf{r}$ is
then computed as: 
\begin{equation}
  \mathbf{r} \;=\; I_{\mathrm{GT}} - I_{\mathrm{scat}}
\label{eq:residual}
\end{equation}
%

In Stage~II, we initialize $\mathcal{G}_{\mathrm{refl}}$ from the
converged $\mathcal{G}_{\mathrm{scat}}$, freeze
$\mathcal{G}_{\mathrm{scat}}$, and optimize only
$\mathcal{G}_{\mathrm{refl}}$.  For each Gaussian disk
$g_i \!\in\! \mathcal{G}_{\mathrm{refl}}$,
we query a learnable
environment map~$E$ along its local reflection~$\omega_{r,i}$ and modulate the result by the Fresnel vector~$\mathbf{f}_{0,i}$:
\begin{equation}
  \mathbf{c}_i^{\mathrm{refl}}
  \;=\;
  \mathbf{f}_{0,i} \;\cdot\; E\!\bigl(\omega_{r,i}\bigr)
\label{eq:per_gauss_color}
\end{equation}
These per-Gaussian colors serve as the color prior for rasterizing the
reflection image~$I_{\mathrm{refl}}$.  
Substituting \cref{eq:residual} into the rendering equation
of~\cref{eq:eq6} and setting the photometric error to zero yields the optimality condition:
\begin{equation}
  R\,\bigl(I_{\mathrm{refl}} - I_{\mathrm{scat}}\bigr)
  \;=\; \mathbf{r}
\label{eq:opt_cond}
\end{equation}
In diffuse regions where $\mathbf{r}\!\approx\!\mathbf{0}$,
\cref{eq:opt_cond} is trivially satisfied by $R\!\to\!0$, reducing
the final image to~$I_{\mathrm{scat}}$.  In reflective regions where
$\lVert\mathbf{r}\rVert$ is large, $R$ must be positive and
$I_{\mathrm{refl}}$ must produce correct view-dependent radiance to
compensate. Notably, satisfying \cref{eq:opt_cond} implicitly forces
the reflection Gaussians to recover accurate surface geometry: because
$E$ is a radiance dictionary indexed by direction, only a correct
$\omega_{r,i}$ retrieves the correct radiance; since $\omega_{o,i}$ is
fixed for a given viewpoint, the only free variable
controlling~$\omega_{r,i}$ is the surface normal~$\mathbf{n}_i$.  The
photometric residuals from Stage~I therefore act as implicit geometric
supervision, driving $\mathcal{G}_{\mathrm{refl}}$ to recover the shape
of reflective and transmissive surfaces without explicit geometry
annotations or priors.

\subsection{Reflection Light Field}
\label{sec:reflmlp}

After reconstructing the geometry of the reflective surfaces,
we seek to recover the reflected near-field environment illumination. To this end, we design a reflection light field, illustrated in
Figure~\ref{fig:refl_mlp} in the Appendix.
Unlike distant illumination that depends only on direction, near-field
reflections vary with both direction and surface position.
The reflection light field therefore takes the reflection
Gaussian's surface position~$\mathbf{x}^{\mathrm{refl}}$ and the
reflection direction~$\omega_r$ as input, enabling the network to
implicitly triangulate the near-field environment across multiple
viewpoints.
The network consists of eight fully connected layers with ReLU
activations and 256 hidden units per layer, followed by a sigmoid
output layer.  Because standard MLPs exhibit a spectral bias toward
low-frequency functions~\cite{rahaman2019spectral}, while specular
reflections often contain high-frequency details, we first lift the
inputs~$\mathbf{x}^{\mathrm{refl}}$ and~$\omega_r$ into a
higher-dimensional space via Fourier feature
encoding~\cite{tancik2020fourier}:
\begin{equation}
\gamma(p)=\bigl(
\sin(2^{0}\pi p),\;\cos(2^{0}\pi p),\;\ldots,\;
\sin(2^{L-1}\pi p),\;\cos(2^{L-1}\pi p)
\bigr)
\label{eq:eq13}
\end{equation}
Before encoding, we normalize $\mathbf{x}^{\mathrm{refl}}$ to
$[-1,1]$ according to the spatial extent of the reflection Gaussian
geometry, and normalize $\omega_r$ to a unit vector.  We set $L=12$
for both $\gamma(\mathbf{x}^{\mathrm{refl}})$ and
$\gamma(\omega_r)$; the relatively high frequency band for the
directional input is chosen to provide sufficient capacity for
modeling the sharp angular variation of specular reflections.
In the first four layers, only the encoded surface
position~$\gamma(\mathbf{x}^{\mathrm{refl}})$ is processed,
rather than being concatenated with the encoded reflected direction
from the outset.  This design allows the network to first extract
position-dependent features such as spatially varying visibility and
local geometric context before fusing them with directional
information, thereby helping disentangle spatially varying properties
from direction-dependent reflectance.  At the fifth layer, the learned
position features are concatenated with the encoded reflected
direction~$\gamma(\omega_r)$, and the encoded surface
position~$\gamma(\mathbf{x}^{\mathrm{refl}})$ is re-injected via a
skip connection.  Because reflected radiance is highly sensitive to
surface location, this skip connection facilitates gradient flow and
preserves fine-grained positional cues, mitigating the
over-smoothing commonly observed in deep
networks~\cite{park2019deepsdf}.  Finally, a sigmoid activation
produces the reflected color.

During the training of the reflection light field, we freeze the
geometry of the reflection Gaussians while allowing the geometry and
appearance of the scattering Gaussians to remain learnable.  This
strategy enables the scattering Gaussians and the reflection light
field to be jointly optimized, thereby achieving accurate scene
reconstruction and inverse rendering of the transmissive scene.

\subsection{Training}
\label{sec:training}

We supervise the rendered image against the ground truth with an RGB
reconstruction loss that combines an $\mathcal{L}_1$ term and a D-SSIM term:
\begin{equation}
  \mathcal{L}_{c}
  = (1-\lambda)\,\bigl\lVert I - I_{\mathrm{GT}}\bigr\rVert_{1}
  + \lambda\,\mathcal{L}_{\text{D-SSIM}}\!\bigl(I,\,I_{\mathrm{GT}}\bigr)
  \label{eq:eq14}
\end{equation}
where $\lambda=0.2$.  To regularize the surface geometry, we adopt
the normal consistency loss $\mathcal{L}_{n}$ from
2DGS~\cite{huang20242d} and apply it independently to both the
reflection Gaussians and the scattering Gaussians:
\begin{equation}
  \mathcal{L}_{n}^{(\cdot)}
  = \sum
    \bigl(1 - \mathbf{n}_{(\cdot)}
              \cdot \hat{\mathbf{n}}_{(\cdot)}\bigr)
  \label{eq:eq15}
\end{equation}
where $(\cdot)\in\{\mathrm{refl},\,\mathrm{scat}\}$,
$\mathbf{n}_{(\cdot)}$ is the rendered normal map obtained by
alpha-blending the normals of the individual 2D Gaussian primitives,
and $\hat{\mathbf{n}}_{(\cdot)}$ is the corresponding normal map
derived from the gradient of the rendered depth map via finite
differences.
While $\mathcal{L}_{c}$ is computed over the entire image, such an
image-wide objective tends to match the average appearance and can
smooth out high-frequency details.  To preserve these details, we
introduce a high-frequency loss $\mathcal{L}_{\mathit{hf}}$ that
focuses the optimization on pixels with the largest rendering errors.
Specifically, let $e(\mathbf{x})$ denote the per-pixel error computed in the same manner as $\mathcal{L}_{c}$,
and let $\tau_{p}$ be the $p$-th percentile of
$\{e(\mathbf{x})\}_{\mathbf{x}\in\Omega}$.  The high-frequency loss
is defined as the mean error over all pixels whose error exceeds
$\tau_{p}$:
\begin{equation}
  \mathcal{L}_{\mathit{hf}}
  = {\textstyle\sum_{\mathbf{x}\in\Omega}
      \mathbb{I}(e(\mathbf{x})>\tau_{p})\,e(\mathbf{x})}
    \;\Big/\;
    \bigl({\textstyle\sum_{\mathbf{x}\in\Omega}
      \mathbb{I}(e(\mathbf{x})>\tau_{p})}+\epsilon\bigr)
  \label{eq:eq16}
\end{equation}
where $\Omega$ is the set of valid pixels, $\mathbb{I}(\cdot)$ is the
indicator function, and $\epsilon$ is a small constant for numerical
stability.  
%
The overall training objective is:
\begin{equation}
  \mathcal{L}
  = \mathcal{L}_{c}
  + \lambda_{n}^{\mathrm{refl}}\,\mathcal{L}_{n}^{\mathrm{refl}}
  + \lambda_{n}^{\mathrm{scat}}\,\mathcal{L}_{n}^{\mathrm{scat}}
  + \lambda_{\mathit{hf}}\,\mathcal{L}_{\mathit{hf}}
  \label{eq:eq17}
\end{equation}

\begin{figure}[t]

  \centering
  \includegraphics[width=\textwidth]{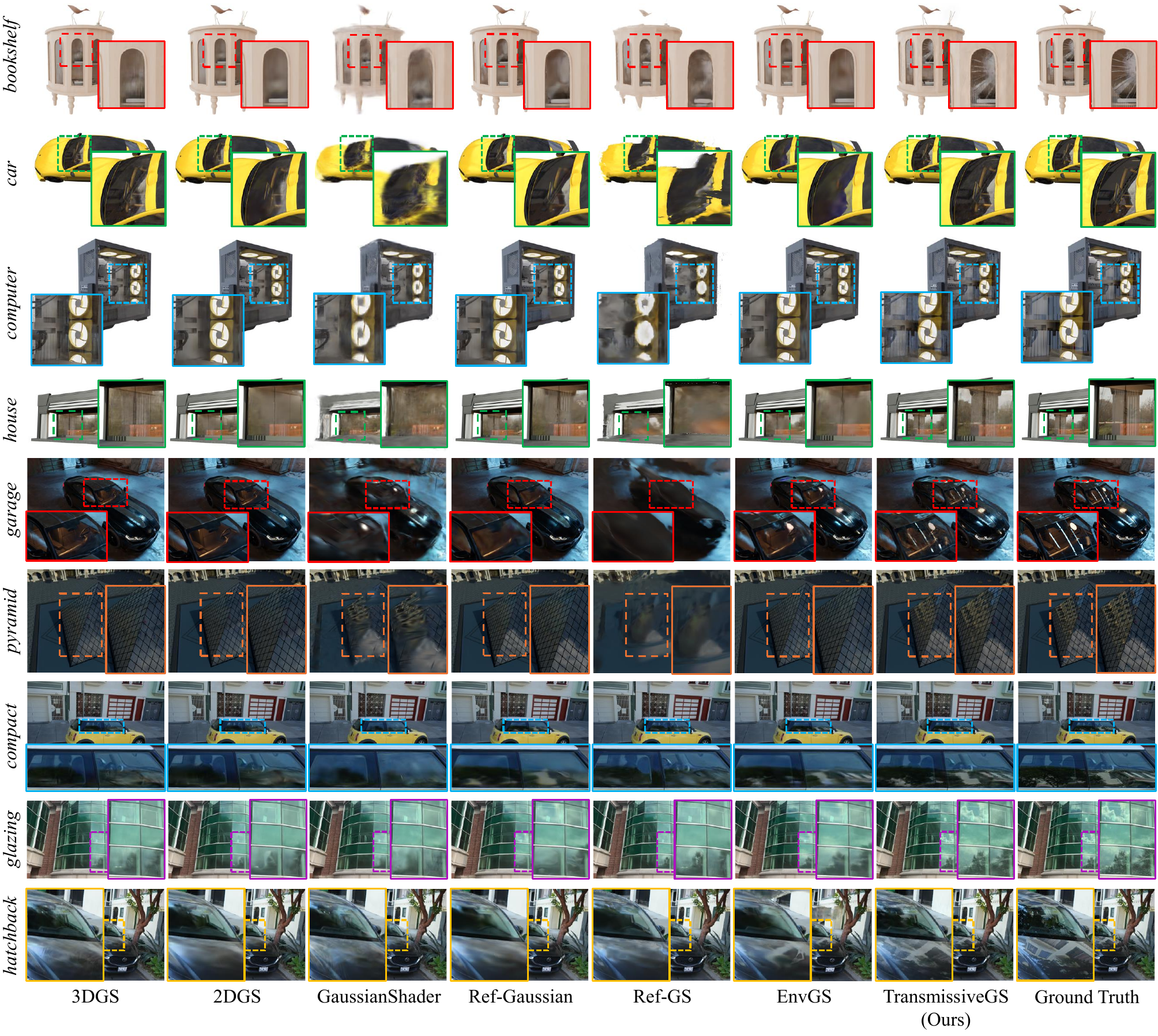}
  \vspace{-0.2in}
\caption{Qualitative comparison of photorealistic rendering quality.
  }
  \label{fig:color_qualitative}

\end{figure}

\section{Experiments}
\label{sec:experiment}

\subsection{Datasets, Evaluation Metrics and Implementation Details}

We evaluate the TransmissiveGS on both synthetic and real-world data.
For synthetic experiments, we construct a new Blender dataset
comprising seven scenes specifically designed for transmissive
scenarios. Among them, \textit{bookshelf}, \textit{car},
\textit{computer}, and \textit{house} each contain a single
transmissive object whose reflected illumination is driven by an HDRI
environment map. The remaining three scenes, \textit{garage},
\textit{pyramid}, and \textit{dome}, include both a transmissive
object and surrounding scene geometry, where reflections arise from
near-field illumination produced by nearby objects. For real-world
experiments, we adopt \textit{compact} and \textit{hatchback} from the
NeRF-Casting dataset~\cite{verbin2024nerf}, and additionally capture
two new scenes (\textit{facade} and \textit{glazing}) that feature transmissive glass curtain walls.

We use consistent training/testing splits across all datasets.
Photorealistic rendering quality is evaluated on all seven synthetic
and four real-world scenes using three standard metrics: PSNR,
SSIM~\cite{wang2004image}, and LPIPS~\cite{zhang2018unreasonable}.
To further assess the geometric accuracy of the reconstructed
transmissive surfaces, we additionally compute the Mean Angular Error in degrees (MAE$^\circ$)~\cite{fouhey2013data} over the transmissive surface regions in the
seven synthetic scenes.

\begin{table}[!t]
  \caption{Quantitative comparison of photorealistic rendering quality.}
  \label{tab:baseline_blender}
\centering
\resizebox{\textwidth}{!}{%
\begin{tabular}{c|ccccccc|cccc}
\toprule
\multirow{2}{*}{\textbf{Scenes}}
& \multicolumn{7}{c|}{\textbf{Blender Synthesis Scenes}}
& \multicolumn{4}{c}{\textbf{Real World Scenes}} \\
\cmidrule(lr){2-8} \cmidrule(lr){9-12}
& \textbf{bookshelf} & \textbf{car} & \textbf{computer} & \textbf{house}
& \textbf{garage} & \textbf{pyramid} & \textbf{dome}
& \textbf{compact} & \textbf{facade} & \textbf{glazing}
& \textbf{hatchback} \\
\midrule


\multicolumn{12}{c}{\textbf{PSNR$\uparrow$}} \\
\midrule
3DGS \cite{kerbl20233d}
  & \second{36.21} & 32.84 & 35.48 & \third{33.55} & 23.56 & \third{32.81}
  & \third{20.80} & \second{29.52} & 23.88 & 27.70 & \third{26.50} \\
2DGS \cite{huang20242d}
  & \third{35.76} & 32.56 & 35.25 & 33.08 & 23.23 & 31.95
  & 20.28 & \third{29.43} & \third{24.44} & 26.60 & 26.43 \\
GaussianShader \cite{jiang2024gaussianshader}
  & 28.51 & 16.88 & 22.88 & 17.12 & 17.09 & 19.72
  & 13.60 & 28.37 & 22.60 & 27.68 & 25.72 \\
Ref-Gaussian \cite{yao2025reflective}
  & 35.64 & \second{34.68} & \third{35.55} & 31.73 & \third{23.58} & \second{33.09}
  & \second{20.94} & 28.83 & 22.76 & \third{27.73} & 26.24 \\
Ref-GS \cite{zhang2025ref}
  & 27.51 & 17.60 & 22.33 & 16.84 & 17.79 & 24.08
  & 14.73 & 29.42 & \second{24.51} & 27.26 & \second{26.54} \\
EnvGS \cite{xie2025envgs}
  & 34.79 & \third{32.92} & \second{35.83} & \second{33.62} & \second{23.64} & 31.82
  & 20.44 & 27.64 & 24.22 & \second{28.92} & 23.74 \\
\textbf{\textit{TransmissiveGS}}
  & \best{37.19} & \best{34.74} & \best{36.93} & \best{34.31} & \best{24.35} & \best{33.49}
  & \best{21.22} & \best{29.65} & \best{24.69} & \best{29.16} & \best{26.79} \\
\midrule


\multicolumn{12}{c}{\textbf{SSIM$\uparrow$}} \\
\midrule
3DGS \cite{kerbl20233d}
  & \second{0.983} & \third{0.963} & 0.976 & 0.958 & 0.730 & \third{0.940}
  & \second{0.721} & \second{0.889} & 0.896 & 0.910 & \second{0.845} \\
2DGS \cite{huang20242d}
  & 0.981 & \second{0.966} & \third{0.980} & \third{0.961} & 0.729 & 0.932
  & 0.678 & \third{0.887} & \second{0.901} & 0.902 & 0.843 \\
GaussianShader \cite{jiang2024gaussianshader}
  & 0.946 & 0.854 & 0.917 & 0.767 & 0.517 & 0.531
  & 0.214 & 0.882 & 0.841 & 0.909 & 0.837 \\
Ref-Gaussian \cite{yao2025reflective}
  & 0.980 & 0.961 & 0.978 & \second{0.966} & \third{0.735} & \second{0.943}
  & \third{0.720} & 0.869 & 0.863 & 0.913 & 0.827 \\
Ref-GS \cite{zhang2025ref}
  & 0.944 & 0.865 & 0.916 & 0.787 & 0.562 & 0.588
  & 0.245 & \third{}{0.887} & \third{0.899} & \third{0.916} & \second{0.845} \\
EnvGS \cite{xie2025envgs}
  & \third{0.982} & 0.958 & \second{0.981} & 0.953 & \second{0.739} & 0.937
  & 0.697 & 0.825 & 0.878 & \best{0.919} & 0.764 \\
\textbf{\textit{TransmissiveGS}}
  & \best{0.985} & \best{0.970} & \best{0.983} & \best{0.971} & \best{0.746} & \best{0.946}
  & \best{0.731} & \best{0.891} & \best{0.904} & \second{0.917} & \best{0.852} \\
\midrule


\multicolumn{12}{c}{\textbf{LPIPS$\downarrow$}} \\
\midrule
3DGS \cite{kerbl20233d}
  & \second{0.031} & \second{0.035} & \third{0.025} & \second{0.035} & \third{0.439} & \second{0.076}
  & \second{0.244} & \second{0.164} & 0.089 & \third{0.089} & 0.201 \\
2DGS \cite{huang20242d}
  & \third{0.033} & 0.041 & \best{0.022} & \third{0.039} & 0.447 & 0.092
  & 0.281 & 0.179 & \second{0.087} & 0.108 & 0.210 \\
GaussianShader \cite{jiang2024gaussianshader}
  & 0.092 & 0.182 & 0.113 & 0.257 & 0.687 & 0.584
  & 0.700 & \third{0.168} & 0.131 & 0.099 & \second{0.197} \\
Ref-Gaussian \cite{yao2025reflective}
  & 0.036 & \third{0.038} & 0.027 & 0.043 & \second{0.431} & 0.082
  & \third{0.246} & 0.206 & 0.111 & 0.098 & 0.231 \\
Ref-GS \cite{zhang2025ref}
  & 0.094 & 0.182 & 0.123 & 0.256 & 0.602 & 0.551
  & 0.707 & 0.169 & \third{0.088} & 0.091 & \best{0.196} \\
EnvGS \cite{xie2025envgs}
  & 0.048 & 0.051 & 0.031 & 0.053 & 0.449 & \third{0.078}
  & 0.263 & 0.240 & \third{0.088} & \second{0.084} & 0.289 \\
\textbf{\textit{TransmissiveGS}}
  & \best{0.029} & \best{0.033} & \best{0.022} & \best{0.033} & \best{0.418} & \best{0.072}
  & \best{0.218} & \best{0.158} & \best{0.085} & \best{0.081} & \second{0.197} \\
\bottomrule
\end{tabular}%
}
\end{table}

\begin{figure}[h]
  \centering
  \includegraphics[width=\textwidth]{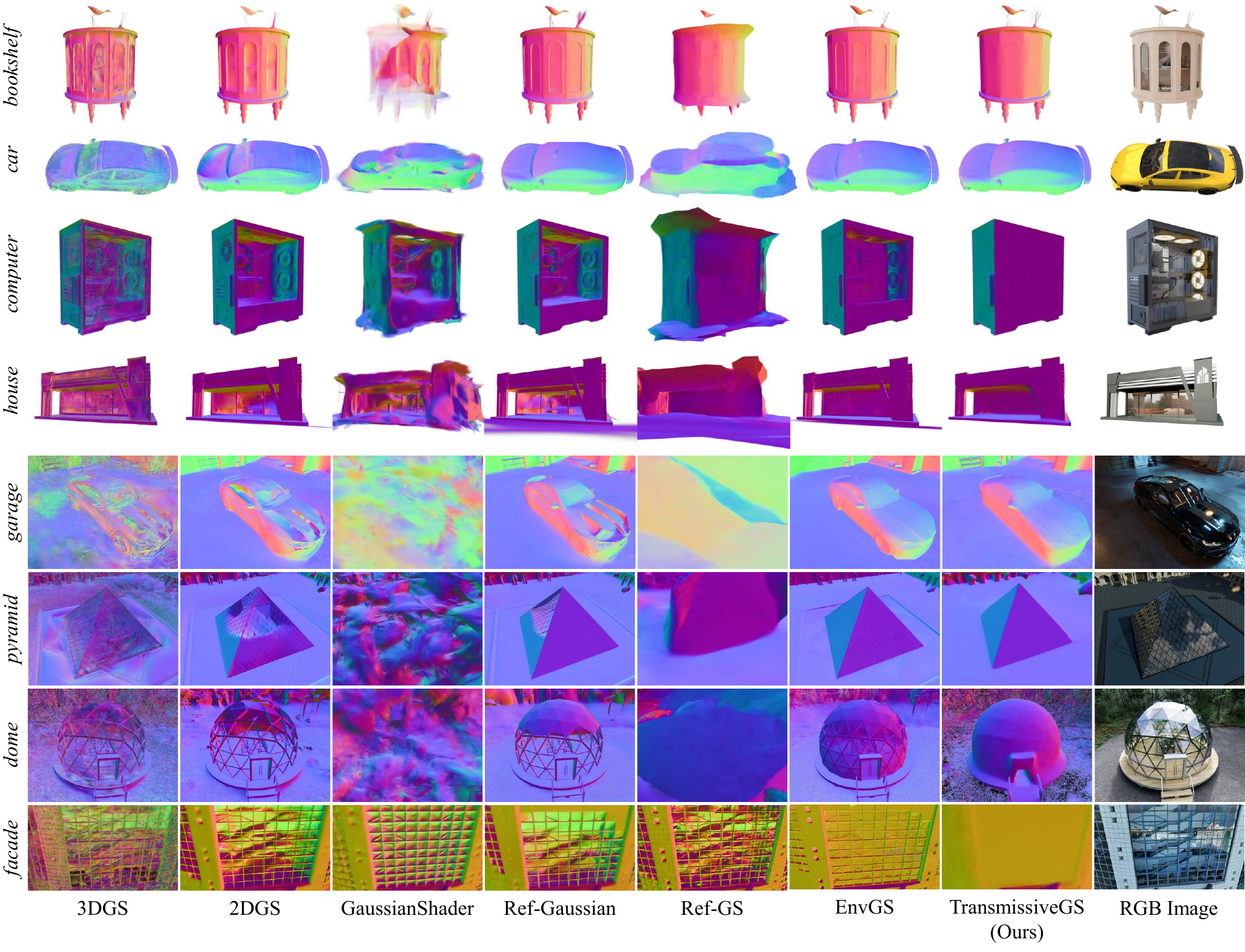}
  \vspace{-0.2in}
  \caption{{ Qualitative comparison of transmissive surface reconstruction.} 
  }
  \label{fig:normal_qualitative}

\end{figure}


All experiments are conducted on a single NVIDIA RTX 4090 GPU with
24\,GB of VRAM. The optimizable parameters comprise those of the
reflection light field, the reflection Gaussians, and the scattering
Gaussians. 
All parameters
are optimized with the Adam optimizer through differentiable splatting
and gradient-based backpropagation. Before being fed into the
reflection light field, surface positions are normalized according to
the spatial extent of the reflection Gaussians.

\subsection{Baseline Comparisons}

We compare our method against current state-of-the-art Gaussian Splatting-based approaches, including GaussianShader \cite{jiang2024gaussianshader}, Ref-Gaussian \cite{yao2025reflective}, Ref-GS \cite{zhang2025ref}, and EnvGS \cite{xie2025envgs}, as well as vanilla 3DGS \cite{kerbl20233d} and 2DGS \cite{huang20242d}. We train all baselines using their publicly available code and configurations.

The quantitative and qualitative results on photorealistic rendering
quality are presented in Table~\ref{tab:baseline_blender} and
Figure~\ref{fig:color_qualitative}. As shown in
Table~\ref{tab:baseline_blender}, our TransmissiveGS achieves a
clear improvement over all baselines. The
qualitative comparisons in Figure~\ref{fig:color_qualitative} further
corroborate this advance. On the
one hand, our method faithfully reconstructs high-frequency near-field
reflections with sharp appearance details, such as the reflected
surrounding structures in \emph{pyramid}, the reflected trees in
\emph{glazing}, and the reflected neighboring building on the car hood
in \emph{hatchback}, as well as local inter-reflections (e.g., the
columns of the house itself reflected on the glass in \emph{house}).
On the other hand, TransmissiveGS simultaneously preserves the
transmitted content behind the transparent surface with high fidelity:
in \emph{compact}, both the window reflections and the interior
details (e.g., the seat and steering wheel) remain clearly visible; in
\emph{hatchback}, the windshield reflections and the interior objects
coexist without mutual degradation. 


In addition, the quantitative and qualitative results of transmissive
surface reconstruction are shown in
Table~\ref{tab:normal_accuracy} and
Figure~\ref{fig:normal_qualitative}. Since EnvGS~\cite{xie2025envgs} leverages
StableNormal~\cite{ye2024stablenormal} as a geometric prior, we exclude it from the quantitative
comparison for fairness. As shown in Table~\ref{tab:normal_accuracy},
TransmissiveGS substantially outperforms all methods across
every scene. The qualitative results in
Figure~\ref{fig:normal_qualitative} further demonstrate that our
method can faithfully recover the geometry of fully transparent
transmissive surfaces. In contrast, although EnvGS incorporates an
external geometric prior, it still fails to reconstruct the
transmissive surface geometry in several challenging cases such as
\emph{computer}, \emph{dome}, and \emph{facade}.

\begin{table}[t]
  \centering
  \normalsize

  \caption{Quantitative comparison of transmissive
  surface reconstruction (MAE$^\circ$).}
  \label{tab:normal_accuracy}

  \small
  \setlength{\tabcolsep}{2.5pt}

  \begin{tabular}{c|ccccccc|c}
    \toprule
    \textbf{Scenes}
      & \textbf{bookshelf}
      & \textbf{car}
      & \textbf{computer}
      & \textbf{house}
      & \textbf{garage}
      & \textbf{pyramid}
      & \textbf{dome}
      & \textbf{Average} \\
    \midrule
    3DGS \cite{kerbl20233d}
      & 56.41 & 41.45 & 48.80 & 46.46
      & 42.15 & 47.35 & 51.64 & 47.75 \\

    2DGS \cite{huang20242d}
      & 37.59 & \third{19.98} & 43.89 & 59.61
      & \third{33.93} & \third{33.41} & 56.94 & 40.76 \\

    GaussianShader \cite{jiang2024gaussianshader}
      & 44.43 & 30.96 & 42.51 & 55.08
      & 45.29 & 45.03 & 47.61 & 44.42 \\

    Ref-Gaussian \cite{yao2025reflective}
      & \third{13.39} & \second{5.91}
      & \third{42.09} & \third{44.05}
      & \second{28.76} & \second{17.57}
      & \third{39.89} & \second{27.38} \\

    Ref-GS \cite{zhang2025ref}
      & \second{8.10} & 23.95
      & \second{22.83} & \second{20.52}
      & 49.63 & 42.42
      & \second{30.59} & \third{28.29} \\

    \textbf{\textit{TransmissiveGS}}
      & \best{7.38} & \best{5.50}
      & \best{4.57} & \best{12.72}
      & \best{7.01} & \best{6.53}
      & \best{8.50} & \best{7.46} \\
    \bottomrule
  \end{tabular}
\end{table}


\subsection{Ablation Studies}

\begin{figure}[t]
  \centering
  \includegraphics[width=\textwidth]{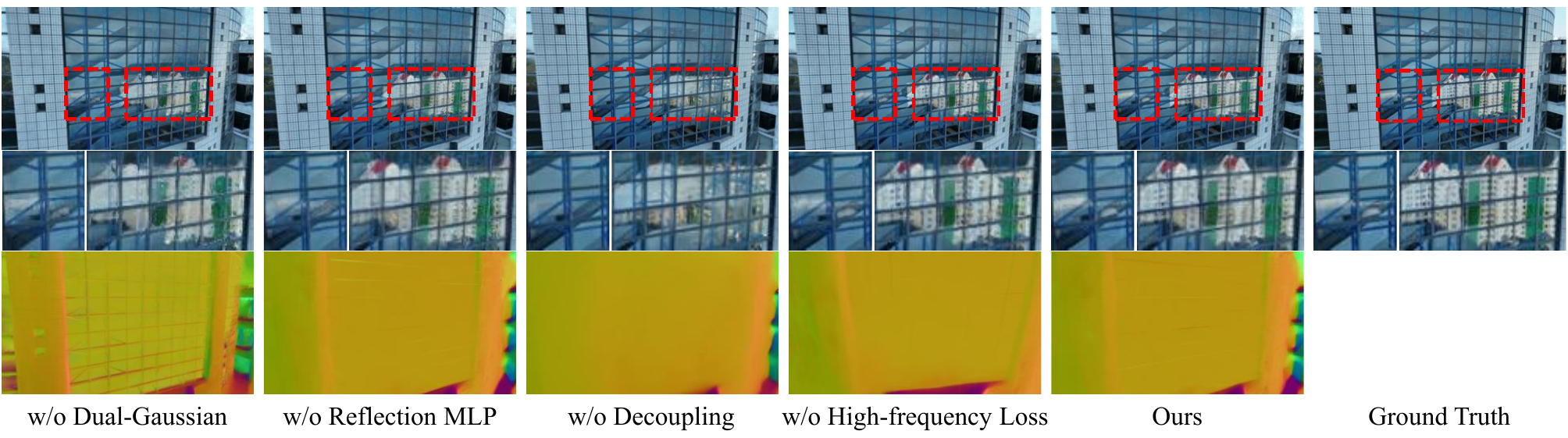}
  \vspace{-0.2in}
  \caption{{ Qualitative results of ablation studies.} 
  }
  \label{fig:ablation}
\end{figure}

We conduct ablation studies on the \emph{facade} scene to validate
the effectiveness and necessity of each design component. The results
are summarized in Table~\ref{tab:ablation} and
Figure~\ref{fig:ablation}.

The \textbf{``w/o Dual-Gaussian''} variant removes the scattering
Gaussians and performs training using only the reflection Gaussians.
As shown in Figure~\ref{fig:ablation}, this variant fails to
reconstruct the transmissive surface geometry, which in
turn degrades the estimation of the reflection light field.

The \textbf{``w/o Reflection MLP''} variant replaces the reflection
light field in Section~\ref{sec:reflmlp} with an environment map-based
reflection formulation. This variant struggles to reconstruct sharp,
high-frequency near-field reflections and introduces parallax errors.

The \textbf{``w/o Decoupling''} variant removes the attributes freezing
of scattering Gaussians during Stage~II and geometry freezing of reflection Gaussians during Stage~III. Without this decoupled optimization strategy, the
two Gaussian sets compete with each other during joint optimization,
leading to degraded transmissive surface geometry and inferior
reflection estimation.

The \textbf{``w/o High-frequency Loss''} variant removes the
high-frequency loss term in Section~\ref{sec:training} during optimization. As a result, the fine
details of the reconstructed reflections are degraded;
e.g., the small reflected structures on the left side of the glass
become blurred.

\begin{table}[!htbp]
  \centering
  \normalsize

  \caption{Quantitative results of ablation studies.}
  \label{tab:ablation}

  \small
  \setlength{\tabcolsep}{6pt}

  \begin{tabular}{lccc}
    \toprule
      & \textbf{PSNR$\uparrow$}
      & \textbf{SSIM$\uparrow$}
      & \textbf{LPIPS$\downarrow$} \\
    \midrule
    w/o Dual-Gaussian
      & 22.98 & 0.827 & 0.152 \\

    w/o Reflection MLP
      & \third{24.20} & 0.858 & 0.124 \\

    w/o Decoupling
      & 23.23 & \third{0.878} & \third{0.100} \\

    w/o High-frequency Loss
      & \second{24.52} & \second{0.891} & \second{0.089} \\
    \midrule
    Ours
      & \best{24.69} & \best{0.904} & \best{0.085} \\
    \bottomrule
  \end{tabular}
\end{table}

\section{Conclusion}
\label{sec:conclusion}

We have presented \textbf{TransmissiveGS}, a novel Gaussian Splatting
framework for accurate reconstruction and photorealistic rendering of
transmissive scenes. Our key contributions include: (i)~a
dual-Gaussian representation that disentangles reflected and
transmitted radiance; (ii)~a residual-guided disentanglement strategy
that enables effective reconstruction of transmissive surface
geometry; and (iii)~a reflection light field formulation that
faithfully captures high-frequency, near-field reflections.
Extensive experiments demonstrate that TransmissiveGS significantly
outperforms existing methods in both inverse rendering quality and transmissive surface reconstruction accuracy. A current limitation is that our method assumes both the reflected and transmitted scene content to be static. In future work, we plan to extend TransmissiveGS to handle dynamic
reflections and transmissions.

\begin{ack}
The authors would like to acknowledge the support by RGC Theme-based Research Schemes (2023/24 T22-606/23-R), RGC Theme-based Research Schemes (2024/25 T22-607/24-N) and Strategic Public Policy Research Funding Scheme (S2024.A7.022). The authors declare that they have no known competing financial interests or personal relationships that could have appeared to influence the work reported in this paper.
\end{ack}

\newpage
\appendix

\section{Architecture of Reflection Light Field.}
\begin{figure}[H]
  \centering
  \includegraphics[
    width=\linewidth,
    height=0.22\textheight,
    keepaspectratio
  ]{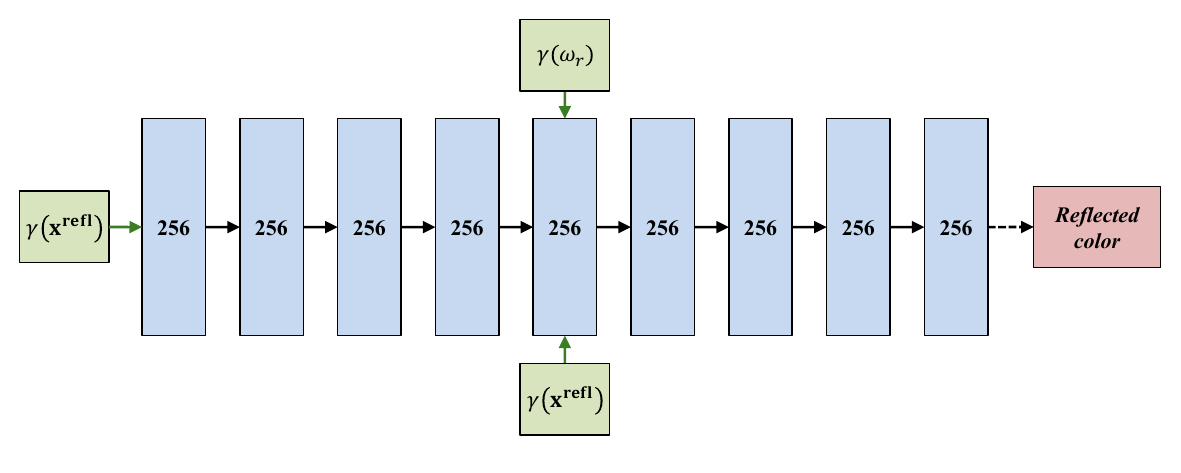}
  \caption{{\bf Architecture of reflection light field.} Green blocks denote input features, blue blocks denote hidden layers, and the red block denotes the output. Solid green arrows indicate data flow. Solid black arrows indicate ReLU activations, and dashed black arrows indicate a sigmoid activation.
  }
  \label{fig:refl_mlp}
\end{figure}

\section{Additional Ablation Studies on Reflection Light Field.}
We further conduct ablation studies on the reflection light field to validate the design choices of its architecture. We investigate the following factors that affect its performance: (1) the network width {\it hidden dim}; (2) the number of layers {\it depth}; (3) the order of Fourier feature encoding {\it L}; (4) whether to re-inject the encoded surface position at intermediate layers via a {\it skip connection}; and (5) whether to input the encoded surface position and encoded reflected direction separately using {\it staged fusion}. We perform these architecture ablations on the {\it garage} case, and present the results in Table~\ref{tab:MLP_abla}.

The results show that removing staged fusion causes a noticeable drop in performance, suggesting that it is beneficial to first let the reflection light field learn position-dependent characteristics and then infer direction-dependent reflections. Incorporating a skip connection consistently improves reflection estimation. Reducing either the network depth or width degrades performance. For the Fourier feature encoding, a sufficiently large order is required for the MLP to model high-frequency reflections; however, overly large orders tend to introduce high-frequency noise and can hurt performance. Overall, these ablation studies confirm that our reflection light field design is effective and reasonable.

\begin{table}[H]
\centering
\caption{Ablation study on the reflection light field.}
\label{tab:MLP_abla}
\renewcommand{\tabularxcolumn}[1]{m{#1}}
\begin{tabularx}{\linewidth}{
  >{\raggedright\arraybackslash}X
  >{\centering\arraybackslash}m{1.4cm}
  >{\centering\arraybackslash}m{1.4cm}
  >{\centering\arraybackslash}m{1.4cm}
}
\toprule
 & \textbf{PSNR$\uparrow$} & \textbf{SSIM$\uparrow$} & \textbf{LPIPS$\downarrow$} \\
\midrule
hidden dim 128, depth 8, L 12, w/ skip connection,

w/ staged fusion
& \second{24.26} & \second{0.744} & 0.424 \\
hidden dim 256, depth 8, L 10, w/ skip connection,

w/ staged fusion
& 24.24 & 0.742 & 0.425 \\
hidden dim 256, depth 8, L 14, w/ skip connection,

w/ staged fusion
& \second{24.26} & \third{0.743} & 0.423 \\
hidden dim 256, depth 6, L 12, w/ skip connection,

w/ staged fusion
& 24.05 & 0.741 & \second{0.420} \\
hidden dim 256, depth 8, L 12, w/o skip connection,

w/ staged fusion
& 24.17 & 0.742 & \third{0.421} \\
hidden dim 256, depth 8, L 12, w/ skip connection,

w/o staged fusion
& 23.96 & 0.739 & 0.424 \\
\midrule
(Ours) hidden dim 256, depth 8, L 12, w/ skip connection, w/ staged fusion
& \best{24.35} & \best{0.746} & \best{0.418} \\
\bottomrule
\end{tabularx}
\end{table}




\section{Failure Mode}

\begin{figure}[h]
  \centering
  \includegraphics[width=0.7\textwidth]{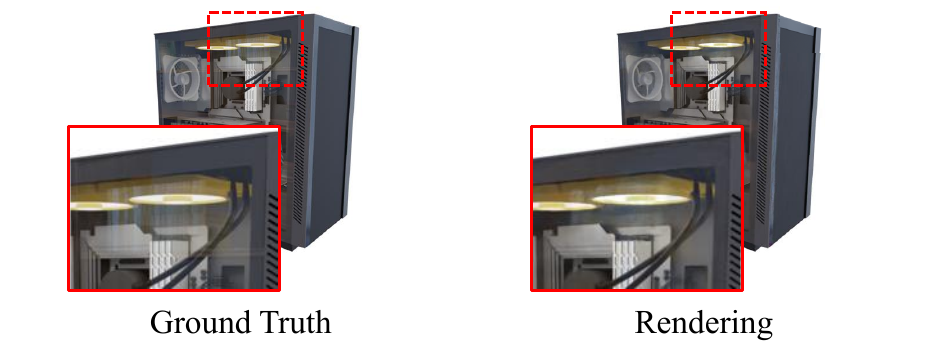}
\caption{Failure mode of our method.}
  \label{fig:failure}
\end{figure}

Figure~\ref{fig:failure} illustrates a failure case of our method. When the transmissive surface is dominated by transmission with only faint reflections at certain viewpoints, the reflection component at these regions can be difficult to estimate accurately and may degenerate.

\newpage
\bibliographystyle{plain}
\bibliography{neurips_2026}


\end{document}